\documentclass[preprint,12pt]{elsarticle}

\usepackage{graphicx}
\usepackage{amssymb}
\usepackage{amsmath}
\usepackage{hyperref}
\usepackage{subcaption} 
\usepackage{lineno}

\usepackage{graphicx}
\usepackage{framed}
\usepackage[margin=3.5cm]{geometry}

\usepackage{tabto}
\usepackage{enumitem}

\makeatletter
\def\ps@pprintTitle{%
 \let\@oddhead\@empty
 \let\@evenhead\@empty
 \def\@oddfoot{\centerline{\thepage}}%
 \let\@evenfoot\@oddfoot}
\makeatother

\begin{document}

\begin{frontmatter}

\title{Applying Transfer Learning To Deep Learned Models For EEG Analysis}



\author[label1,label2]{Axel Uran}
\author[label2]{Coert van Gemeren}
\author[label2]{Rosanne van Diepen}
\author[label1]{Ricardo Chavarriaga}
\author[label1]{Jos\'e del R. Mill\'an}
\address[label1]{Defitech Chair in Brain-Machine Interface, Ecole Polytechnique F\'ed\'erale de Lausanne (EPFL)}
\address[label2]{Alpha.One, Weena 250, 3032 AC, Rotterdam, The Netherlands}

\begin{abstract}
The introduction of deep learning and transfer learning techniques in fields such as computer vision allowed a leap forward in the accuracy of image classification tasks. Currently there is only limited use of such techniques in neuroscience. The challenge of using deep learning methods to successfully train models in neuroscience, lies in the complexity of the information that is processed, the availability of data and the cost of producing sufficient high quality annotations. Inspired by its application in computer vision, we introduce transfer learning on electrophysiological data to enable training a model with limited amounts of data. Our method was tested on the dataset of the BCI competition IV 2a and compared to the top results that were obtained using traditional machine learning techniques. Using our DL model we outperform the top result of the competition by 33\%. We also explore transferability of knowledge between trained models over different experiments, called \textit{inter-experimental transfer learning}. This reduces the amount of required data even further and is especially useful when few subjects are available. This method is able to outperform the standard deep learning methods used in the BCI competition IV 2b approaches by 18\%. 
In this project we propose a method that can produce reliable electroencephalography (EEG) signal classification, based on modest amounts of training data through the use of transfer learning. 
\end{abstract}

\end{frontmatter}

\section{Introduction} 
Traditionally, deep learning (DL) methods require large amounts of data to avoid overfitting. Since data collection (i.e. functional Magnetic Resonance Imaging, electroencephalography) is relatively expensive and time-consuming, its application in the field of neuroscience is challenging. Traditional machine learning algorithms do not require as much data and are typically preferred over DL models \cite{Schmidhuber2015DeepOverview}. However, the architecture of the DL model plays an important role on its performances in general \cite{LeCunDeepLearning,Garcia-GarciaASegmentation}. We show that selecting a suitable network design increases performance in a traditional motor imagery task above that of traditional machine learning algorithms, even with limited training data. 

DL methods gained in popularity with AlexNet \cite{KrizhevskyImageNetNetworks} in 2012. This success is explained by the drop in error rate compared to the previous results at the Large Scale Visual Recognition Challenge (LSVRC) \cite{Russakovsky2015ImageNetChallenge}. AlexNet's results popularized the use of DL in a multitude of fields. The method is now applied in a plethora of applications, such as computer vision \cite{Iizuka2016LetColor,CaoRealtime}, medical diagnostics \cite{Gargeya2017AutomatedLearning} and recently brain-computer interface (BCI) \cite{Schirrmeister2017DeepVisualization,Lawhern2016EEGNet:Interfaces}.

The use of DL for BCI is a recent development, until now BCI applications mostly relied on standard machine learning techniques such as Linear Discriminant Analysis (LDA) or Support Vector Machines (SVM), supervised feature selection or on knowledge from the field of neuroscience \cite{Nicolas-Alonso2012BrainReview,Vaid2015EEGReview}. Recently there has been an increase in the use of DL for EEG decoding on different tasks, such as predicting the sex of the user \cite{VanPutten2018PredictingLearning}, epileptic seizure prediction \cite{Hosseini2017Cloud-basedPrediction} or classification of EEG motor imagery signals \cite{Tabar2017ASignals}. 

Although EEG is the most convenient way to record brain activity, there are disadvantages, such as a high intra- and inter-individual variability, a low signal to noise ratio, and high monetary and time costs \cite{Mak2009ClinicalProspects}. To which other constraints, such as the physical type of device that prevent the user for wearing it for long periods and the difficulty to have a long-term controlled recording make the acquisition of labeled EEG datasets an arduous task. This created the incentive to look for solutions which allow for a reduction in the amount of data needed, while keeping model performance at acceptable levels \cite{Abdulkader2015BrainChallenges}. In the context of deep learning, Transfer learning (TL) allows to speed up the rate at which models converge. Features acquired by training on one dataset can be reused to boost convergeance when training another model on a smaller dataset \cite{SuTransferNetwork,Weiss2016ALearning}. TL approaches in the context of BCI, such as spectral transfer learning, are not following the same techniques \textit{per se} \cite{Waytowich2016SpectralInterface}. Here, information geometry is used to perform spectral transfer learning for user-independent BCI applications. Another approach is conditional TL to improve EEG-based emotion classification \cite{Lin2017ImprovingLearning}.

Large inter- and intra-subject variability \cite{Nicolas-Alonso2012BrainReview} make it difficult to reach high classification accuracy. Generalizing to data of subjects not seen during training also proves difficult when only small amounts of training data are available. To obtain a model that generalizes well to unseen data, we will be testing multiple techniques inspired by TL techniques used in computer vision. This will allow us to study the transferability of knowledge between subjects and sessions, which we will call \textit{intra-experimental transfer}. 

We will study the capabilities of a model to transfer knowledge from one experiment settings to another, which we will call \textit{inter-experimental transfer}. Here we will pre-train the model on a first dataset from one experiment and then fine tune it on a second dataset from another experiment.

\subsection*{Related Work}
The most recent DL architecture used in EEG analysis uses convolution in most of their layers \cite{Nurse2016DecodingLearning,Schirrmeister2017DeepPathology}, these type of DL architectures are called Convolution Neural Networks (CNN) \cite{LeCunDeepLearning}.
One use case of DL in the field of EEG analysis is the work by Van Putten et al. \cite{VanPutten2018PredictingLearning}. They show that it is possible to predict the sex of the subject solely based on EEG activity. To train the model, they used 40,000 non overlapping two second epochs of EEG activity while subjects had their eyes closed. Van Putten et al. found that their model is able to predict the sex of the subjects with an accuracy of 81\%. Applying this model to other tasks would require an equally large data set for every subsequent task. This is often not feasible.

Lawhern et al. \cite{Lawhern2016EEGNet:Interfaces} proposed a more compact architecture with four layers, which is able to perform at the same level as most of the traditional methods of EEG analysis with a low number of subjects. They analyzed the impact of the training dataset size on a ERP-based signals decoding task. Performance hits a plateau after 1000 trials which shows that they are able to train the model with low amounts of data, while still getting good results. They use dropout \cite{Srivastava2014Dropout:Overfitting}, which allows it to reduce over-fitting even with a small dataset. In this scenario, Lawhern et al. train the model on data of all but one subject, which is kept out during training for validation and testing. This approach yields impressive results. Here, we extend their work by introducing new learning techniques to an architecture that is comparable to theirs. 

Schirrmeister et al. \cite{Schirrmeister2017DeepVisualization} introduce a model inspired by deep CNNs. Deep CNNs allowed for end-to-end learning in the field of computer vision \cite{KrizhevskyImageNetNetworks,SuTransferNetwork,Shan2017UnsupervisedRegistration}. The motivation of Schirrmeister et al. is to introduce methods like these in EEG analysis. First they present the deep ConvNet, close to the standard architecture used in computer vision. Finally they introduce shallow ConvNets, which are inspired by the \textit{Filter Bank Common Spatial Patterns} (FBCSP)\cite{KaiKengBCIBy} pipeline, specifically tailored to decode band power features in EEG signals. The main difference with the two previous works is the introduction of a technique for data augmentation inspired by the field of computer vision, called \emph{cropped training strategy}. The authors augment the data by creating one crop per time-step in the the EEG trial time series. Following this strategy they adopt a cropping approach that leads to 625 crops per epoch. The motivation of the authors for implementing aggressive cropping on the training data, is to force the model to put emphasis on the features that are present in all crops of the trail. To attain a level that is comparable to the model introduced by Lawhern et al. \cite{Lawhern2016EEGNet:Interfaces}, Schirrmeister et al. had to train the model on a supplementary dataset, called the high-gamma dataset, which sees a four fold increase in datapoints in comparison to the dataset of Lawhern et al. \cite{Lawhern2016EEGNet:Interfaces}. This addition of new data shows that their model requires a large number of training examples, which is not always available.

\textit{Spectral Transfer using Information Geometry} (STIG) introduced by Waytowich et al. \cite{Waytowich2016SpectralInterface} addresses transferability using an unsupervised technique. STIG ranks and combines unlabeled predictions from an ensemble of information geometry classifiers, built on data from individual training subjects. This method outperforms existing calibration-free techniques as well as traditional within-subject calibration techniques when limited data is available. This method, while being an interesting solution to intra- and inter-subject variability, is not comparable to the TL methods that have been applied specifically to CNNs in the field of computer vision.

Yosinski et al. \cite{Yosinski2014HowNetworks} shows that CNNs trained on natural images will exhibit features similar to Gabor filters and color blobs on its lowest layers. These first-layer features are not dataset or task specific and can thus be reused in another context. A commonly used TL technique in computer vision, called fine tuning, is based on this principle. By freezing the lower layers of a trained model, we can reduce the number of parameters that have to be optimized for a specific task, which allows for a lower number of examples during training. Shirrmeister et al. \cite{Schirrmeister2017DeepVisualization} show in their work that their CNNs are able to learn spectral power modulations. Lawhern et al.'s model also shows that the lower layers enable efficient extraction of frequency-specific spatial filters \cite{Lawhern2016EEGNet:Interfacesb}. By combining the findings of the previous research groups, we believe that freezing the lower layers of a CNN model trained for EEG analysis could retain the knowledge of differentiating distinct? different spectral power modulations. This approach, which we call \textit{intra-experimental transferability}, reduces the number of subjects needed to train a model for another task. If successful, it would increase the number of possible applications of BCI using DL. 

We have found similarities related to \textit{intra-experimental transferability} in the work of Sakhavi et al. They showed that it is possible to improve model performance by pre-training a CNN model on a set of subjects and fine tune it for another user \cite{Sakhavi2017ConvolutionalBCI}. The main difference with our method is that they apply a FBCSP-based Envelope Extraction before feeding the data into the model. They consider the envelope energy a feature. This reduces the number of input features and allows a model to train more easily. Our approach is to have an end-to-end learning method, therefore we do not apply envelope extraction to the data.

To expand on the \textit{intra-experimental transfer} we will also perform \textit{inter-experimental transfer}. To our knowledge we are the first to perform this task in the field of EEG decoding.

\section{Methods}\label{sec:methods}

The pipeline created to train the models that are used to classify the EEG data consists of an Input, Output and Processing section. A detailed description of their purpose is given below.  

\textbf{Inputs}\tabto{32mm} The input of the pipeline consists of the \textit{DL Architecture} and \textit{Dataset} modules. The DL Architecture is used to specify different network designs that optimize performance for specific tasks. The dataset module contains a default dataset to which the different architectures are compared. In our project the default is the BCI Competition IV 2a \cite{BrunnerBCIParadigm} dataset.

\begin{figure}
\captionsetup{singlelinecheck=off}
\centering
\includegraphics[width=\textwidth]{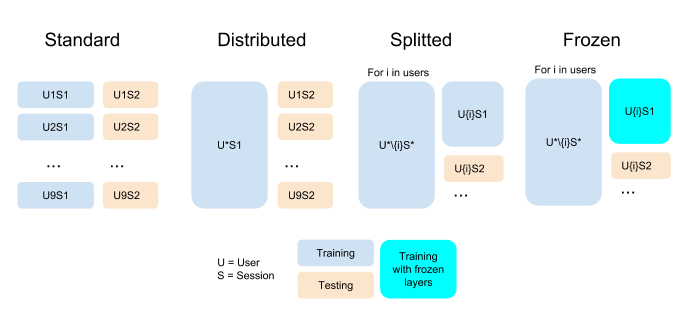}
\caption{Schematic overview of the different training strategies. [Splitting is an irregular verb. The past tense is Split. Please update image]. In blue the data selection procedure for training a model is shown. In yellow we show the data selection procedure for testing an optimized model. We separate the data over \textit{Users} and \textit{Sessions} in four different ways: \textit{Standard Learning}, \textit{Distributed Learning}, \textit{Split Learning} and \textit{Frozen Learning}. We describe each training strategy in detail in Section~\ref{sec:methods}. For \textit{Frozen Learning}, the turquoise block indicates that we retrain the data of specific users without updating part of the model parameters.} 
\label{fig:Trng_met}
\end{figure}

\textbf{Processing}\tabto{32mm} The input data is split into training and testing data. Before training, the parameters of the \textit{Data Channel}, \textit{Data Filter} and \textit{Training Technique Selections} modules were set based on experimental conditions. Subsequently, the model is optimized on the training data and evaluated on the test data. 
The \textit{Data Channel Selection} module allows us to adapt the pipeline easily to the given number of channels. Depending on the settings of the experiments or the dataset, there can be a different number of channels available. The \textit{Data Filter Selection} preprocessing module allows for applying specific filters to the the data. Different tasks rely on different frequency bands. The signal-to-noise ratio could be improved by selecting specific frequency bands that correlate with specific tasks. The \textit{Hyperparameter Selection} module is added to perform automatic hyperparameter selection based on cross-validation on the training data. The \textit{Training Technique Selections} module handles different methods for training. The input data was split over \textit{User} in addition to \textit{Session} subsets, to account for \textit{intra-subject} variability, which is illustrated in Figure~\ref{fig:Trng_met}. Based on the way the data is split into subsets of training and testing data, four optimization strategies can be identified:
\begin{itemize}[leftmargin=!,labelindent=5pt,itemindent=-114pt]
  \setlength\itemsep{0mm}
  \item \textit{Standard Learning}\tabto{0pt}The model is trained on a single subject and tested on another session of the same subject, with data that was not seen during training. This was done for every subject.
  \item \textit{Distributed Learning}\tabto{0pt}The model is trained on data from the first sessions of all subjects. We tested the resulting model on each individual subject, using data from a different session than the one it was trained on.
  \item \textit{Split Learning}\tabto{0pt}The model is trained on data from sessions of all but one user and then retrained on the first session of the last user with the previously initialized parameters. Finally the model is tested on the second session of the same user.
  \item \textit{Frozen Learning}\tabto{0pt}Similar to Split Learning but when the model is retrained, the lower layers will be frozen, thus reducing the amount of parameters to be trained. The number of layers to be frozen can be manually selected.
\end{itemize}

The \textit{Model Training} trains the selected model on the data that has been preprocessed, the hyperparameters that have been selected and the training technique that has been selected in the previous modules. The \textit{Model Selection} allows for a selection of the models produced. 

\vspace{0.5em}

\textbf{Outputs}\tabto{32mm}To be able to evaluate model performance we have two types of output units. The \textit{Model evaluation} outputs multiple metrics: a confusion matrix, a precision recall curve and a loss and learning rate curve. The \textit{Fully Trained Model} saves the model in a reusable format to allow for reproducibility of the model results.

\subsection*{Datasets}
\begin{figure}[h]
\centering
\includegraphics[width=\textwidth]{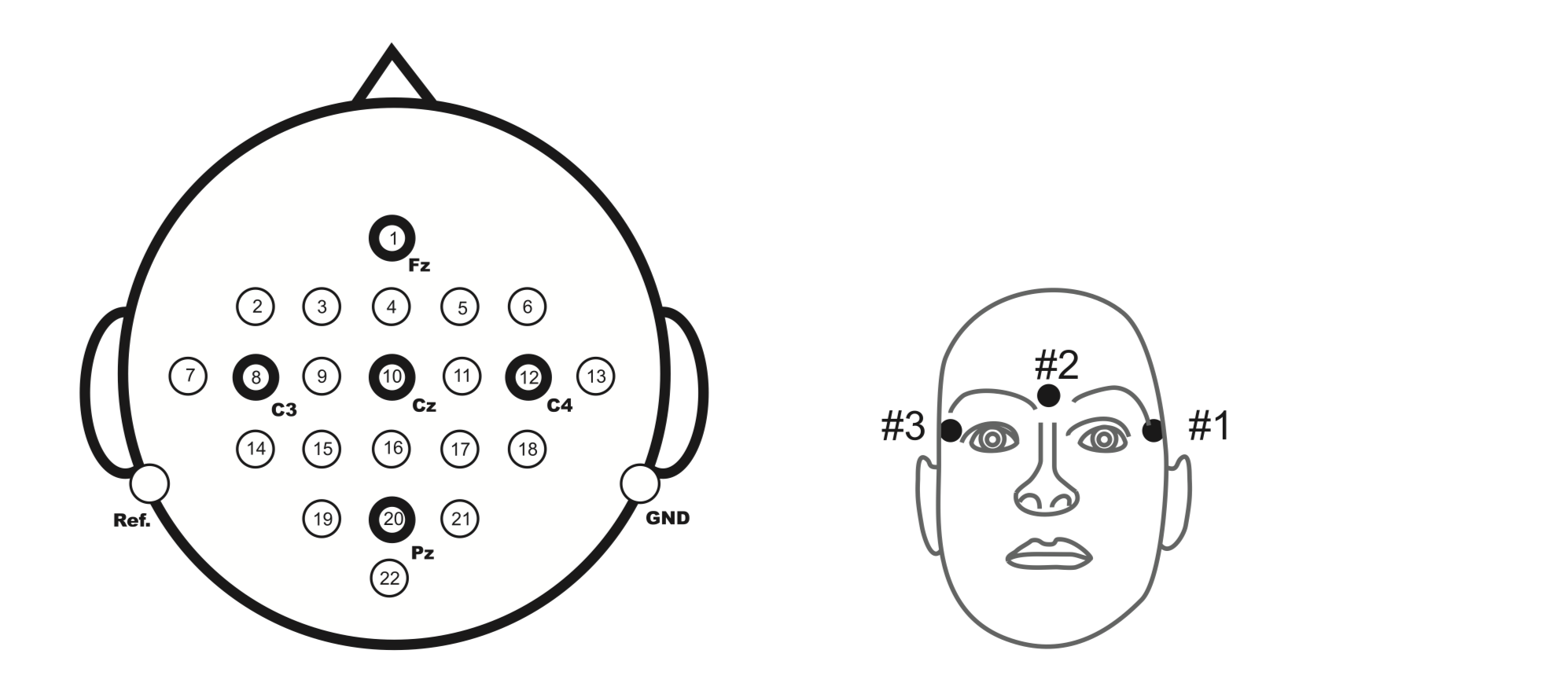}
\caption{Left: Electrode Montage corresponding to the to the international 10-20 system. Right: Electrode montage of three monopolar EOG channels.}
\label{fig:elec_place}
\end{figure}

The data of the BCI competition 2a is composed of nine subjects performing four imagery motor tasks: imagination of the movement of the left hand, right hand, both feet and tongue. The challenge of the competition is to discriminate between these four movement classes. The data is recorded with 22 EEG electrodes and three EOG electrodes placed as in Figure \ref{fig:elec_place}. Two sessions on different days were recorded for each subject, each session is composed of six runs and each run consists of 48 trials which yields a total of 288 trials per session. Due to unavailability of the EOG channels in one of the subjects we reduced the total number of subjects to eight. 

To test the transferability of models we will also work with a second dataset: BCI Competition IV 2b~\cite{LeebBCIParadigm}. The data is composed of nine subjects performing two imagery motor tasks: imagination of the movement of the left hand and of the right hand, which are the two data classes. The data is recorded with three EEG electrodes (C3, Cz and C4 following the 10-20 system) and three EOG electrodes, placed as in Figure \ref{fig:elec_place}. Two sessions were recorded on different days, for each subject, the first session is composed of three runs and the second session is composed of two runs. The first session is composed of 240 trials and the second of 320 trials.

The electrode placement, the sampling rate and the pre-processing were done in a similar fashion as in dataset 2a. 

\subsection*{Inter-experimental Transfer}
A solution to the problem of using deep learning with a low amount of training data could be the reuse of models that have been pre-trained. To test this hypothesis we trained a model on a four movement Motor Imagery (MI) task and then reused the output model on a two movement MI task. We also performed the reverse: we first trained a model on two movements and then transfered it to a four movement MI task. The outcome of these strategies was compared with the intra-experimental learning techniques.

\subsection*{Metrics}
To assess the performance of our models we use multiple metrics that are produced by the Model Evaluation unit. \textit{Accuracy} is used for optimizing the models. A \textit{confusion matrix} and a \textit{precision recall curve} allow us to analyze the tendency of a model to overfit, and to analyze its sensitivity and specificity. We use the $\kappa$ coefficient \cite{Sim2005Number3} to compare our results to the winners of the BCI competition IV 2a \cite{BrunnerBCIParadigm}. The $\kappa$ coefficient shown in Equation \ref{eq:kappa}, is a metric that compensates class imbalance by subtracting the proportion of true positive classifications to be expected by chance ($p_e$), from the accuracy of the tested model ($p_0$). $p_e$ is calculated as the proportion of the most frequently occurring class over the total number of classes we consider. 
%
%

\begin{equation}
\kappa = \frac{p_0 - p_e}{1 - p_e}
\label{eq:kappa}
\end{equation}



\subsection*{Hyperparameter estimation}
To fix the hyperparameter settings a cross-validation was performed on the training data. Performance was tested using the Standard Learning stategy. The hyperparameters were not optimized for each training approach separately, since we want to exclusively compare the learning strategies. Hyperparameters that were optimzed are: the dropout value, the use of filters and the use of channels (in this particular order). 

\subsubsection*{\small Dropout optimization}
Dropout is a technique used in deep learning to prevent overfitting\cite{Srivastava2014Dropout:Overfitting}. This technique will randomly ignore some neurons in the network, the value being the probability of a neuron to be ignored. Reducing the co-dependency amongst specific neurons allows for a more widespread information flow. We optimized the dropout value by evaluating values between 0.0 and 0.9 with an interval of 0.1. A drop-out value of 0.1 yields, based on the median of all the users, the best accuracy and is used in all experiments.

\subsubsection*{\small Data filtering}
Lawhern et al. \cite{Lawhern2016EEGNet:Interfaces} train the ConvNet model directly on the data, without filtering. In contrast, Schirrmeister et al. \cite{Schirrmeister2017DeepVisualization} apply a 4 Hz high pass filter to the data. They justify this by noting that filtering this band removes slow moving artifacts from the signal, that are caused by eye movements. To be able to compare our results to both EEGNet and ConvNet, model performance evaluated with and without low-pass filtering. Filtered data (4 Hz high pass non causal) produced slightly higher accuracy and is therfore used in all experiments.

\subsubsection*{\small Channel selection}
The channel selection module allowed us to perform two experiments: test for training artifacts resulting from recorded eye movement present in the training data and to test the performance of the model with a lower number of channels. In the first case we remove the EOG channels from the training and testing data. For the second case we keep five channels: Fz, C3, Cz, C4 and Pz, circled in bold in Figure \ref{fig:elec_place}. This module will also allow us to compare the recordings between experiments with a different number of input channels. The model that keeps all the channels, including the EOG channels, produces the best results and is used in the experiments.

\subsection*{Experiments}
To evaluate our method and learning techniques, we tested and optimized the EEGnet architecture proposed by Lawhern et al.~\cite{Lawhern2016EEGNet:Interfaces}, using the data from BCI competition IV 2a. Results will be compared to the winners of this competition using the $\kappa$ coefficient.  Finally, in addition to \textit{intra-experimental} transferability (used in Split Learning and Frozen Learning) we tested the possibility of \textit{inter-experimental transferability}.

\subsection*{Training techniques}
Following the EEGNet model optimization for the Standard Learning, all different training strategies will be compared. For the Distributed Learning our assumption is that the EEGNet should be able to generalize well on the data if presented to different subjects during training. For the Split and Frozen Learning our assumption is that we can reuse the features that have been learned on the other subjects. The different training strategies are compared based on their performance with the EEGNet architecture.

\subsection*{Inter-experimental transferability}
To test \textit{inter-experimental transferability} we trained the models on the dataset 2b in a similar fashion as done for the dataset 2a. These results are used as benchmark values to assess the \textit{inter-experimental transfer} performance. Two approaches were then tested: 1) standard transfer learning in which we use the model and only train a second time on the first session of the user in a way similar to Standard Learning 2) split transfer learning in which we perform the Split Learning approach with the pre-trained model. Note that since the number of classes is different we cut out the top classification layer and replaced it with another one that maps to the correct number of classes. Another point to note is that we reduced the number of input channels to three EEG and three EOG to be able to transfer the model between the datasets.


\begin{figure}[h]
\centering
\includegraphics[width=0.8\linewidth]{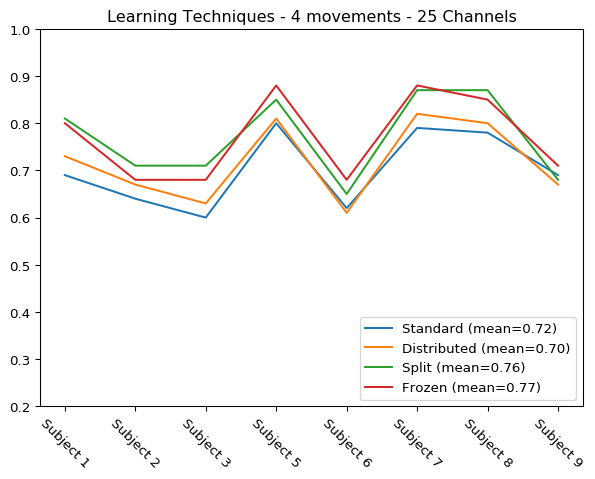}
\caption{Standard vs Distributed vs Split vs Frozen Learning}
\label{fig:mdls}
\end{figure}

\begin{figure}
\begin{subfigure}[h]{0.5\linewidth}
\includegraphics[width=\linewidth]{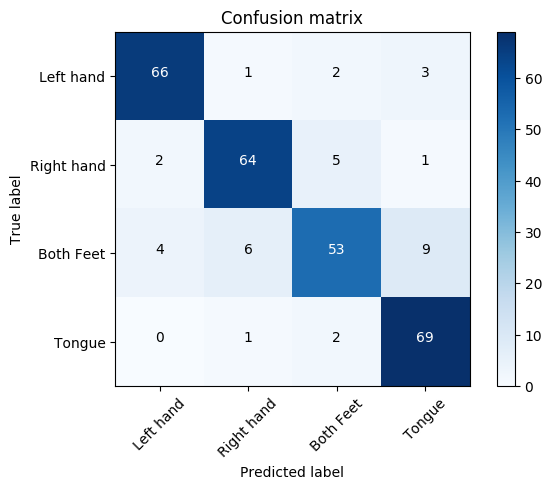}
\end{subfigure}
\hfill
\begin{subfigure}[h]{0.5\linewidth}
\includegraphics[width=\linewidth]{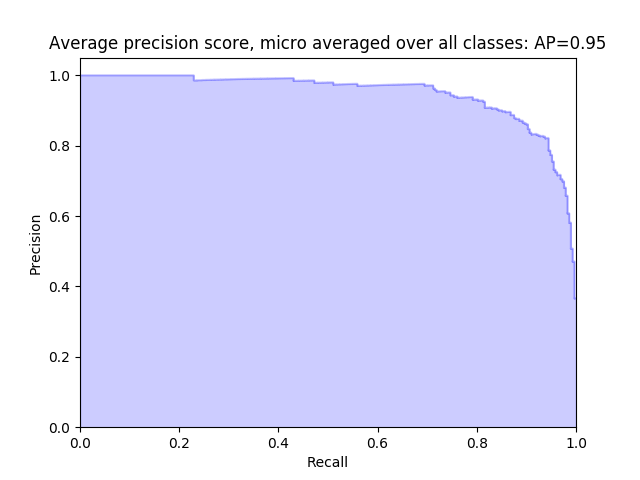}
\end{subfigure}%
\caption{Performance data for the best scoring subject (\#8). Left: Confusion matrix. Right: Precision recall curve.}
\label{fig:Subj8}
\end{figure}

\begin{figure}
\begin{subfigure}[h]{0.5\linewidth}
\includegraphics[width=\linewidth]{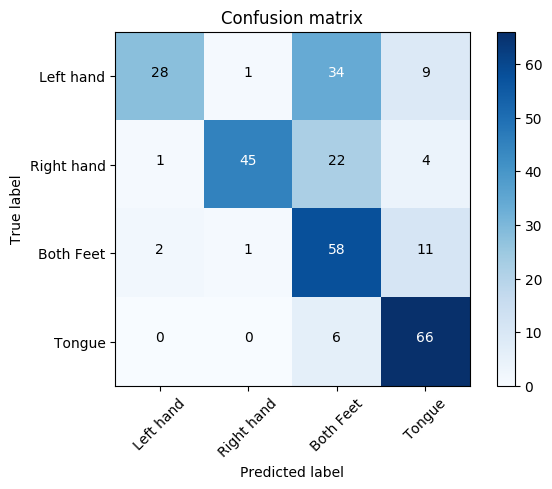}
\end{subfigure}
\hfill
\begin{subfigure}[h]{0.5\linewidth}
\includegraphics[width=\linewidth]{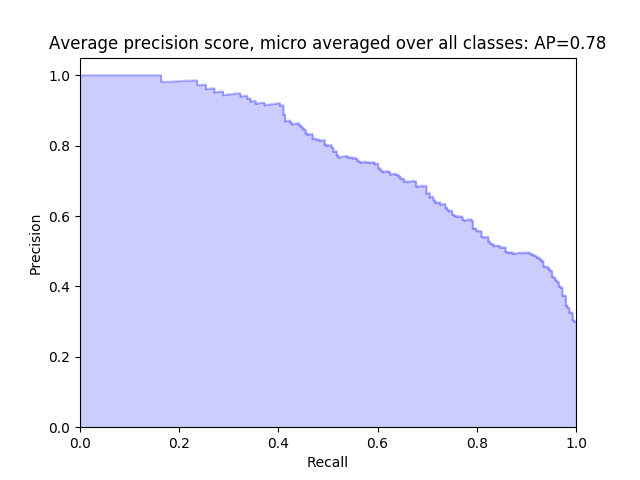}
\end{subfigure}%
\caption{Performance data for the worst scoring subject (\#6). Left: Confusion matrix. Right: Precision recall curve.}
\label{fig:Subj6}
\end{figure}

\section{Results}


\subsection*{Training techniques}
Based on the mean accuracy over the subjects, we order the training strategies from high to low. Frozen Learning performs best (0.770), followed by Split learning (0.768), both having a p-value of 0.13 compared to Distributed learning (0.718) and Standard learning (0.701). Accuracies per subject can be found in Figure \ref{fig:mdls}. Comparing the $\kappa$-score of our best performing model (Frozen Learning: 0.76) to the best result of the winner of the BCI competition IV 2a (0.57), we increase performance by 33\%.

Model evaluation metrics are presented for the best (Figure \ref{fig:Subj8}) and worst (Figure \ref{fig:Subj6}) performing subjects of the model trained by the Frozen Learning strategy. In the case of the worst scoring subject the left and right hand are often confused with the movement of both feet, as can be seen in the confusion matrix of Figure \ref{fig:Subj6} where Left hand and Right hand labels are confused 34\% and 22\%, with the Both Feet label. The same pattern was found for the other learning strategies.

\subsection*{Inter-experimental transferability}

\begin{figure}
\begin{subfigure}[h]{0.5\linewidth}
\includegraphics[width=\linewidth]{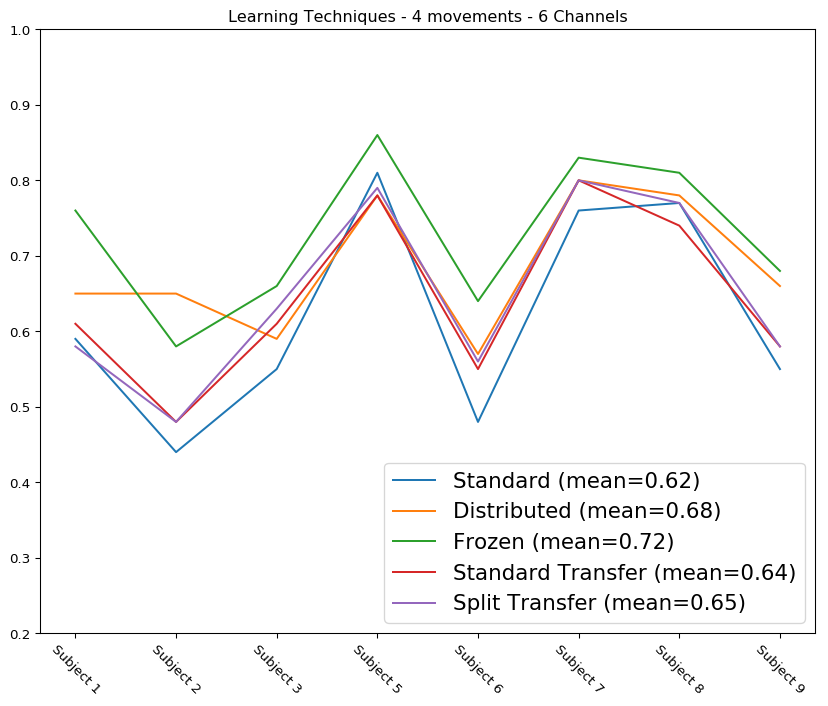}
\end{subfigure}
\hfill
\begin{subfigure}[h]{0.5\linewidth}
\includegraphics[width=\linewidth]{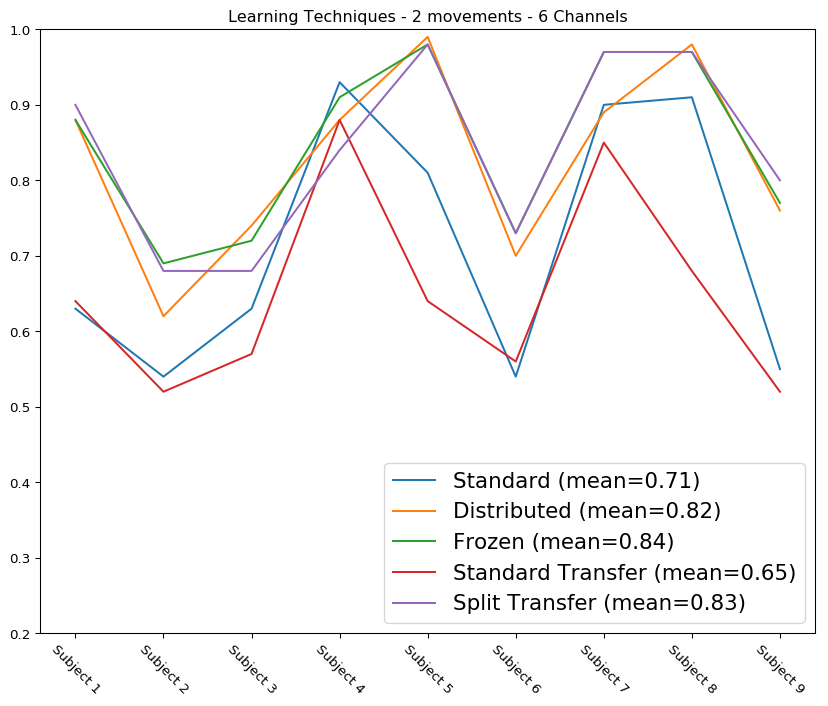}
\end{subfigure}%
\caption{Left: Accuracies of different learning methods - Dataset 2a; Right: Accuracies of different learning methods - Dataset 2b}
\label{fig:transfer}
\end{figure}

As can be seen in Figure \ref{fig:transfer}, the standard transfer method performs on average 5\% better than the standard learning in the transfer to four-movement task but 6\% worse in the transfer to two movement task. The Split transfer method performs better overall than Standard transfer learning, 6\% in the case of four movement task and 18\% increase in the case of two movement task. In both cases the transfer models performs worse than the Frozen and Distributed learning but note that the Split transfer method in the case of two movements is performing around the same level as Frozen and Distributed Learning. 

\section{Discussion}
The goal of the current investigation was to determine whether a model created using deep learning could improve performance on a classification task trained with limited amounts of neurophysiological data. With a carefully chosen architecture we show that despite training on a low amount of data, we can improve over traditional machine learning techniques with an increase in $\kappa$-score of 33\%.

The Frozen Learning model performs best, based on accuracy and $\kappa$-score, closely followed by the Split Learning model. We explain the high performance of both learning strategies through the use of information transfer between subjects (i.e. intra-experimental transfer) and between experiments (i.e. inter-experimental transfer). These neural network models, inspired by Computer Vision, learn generic feature representations that occur in nearly all training samples available, at the lower model layers. At higher layers they learn more specific feature representations, that are tailored to a specific subject or experimental task. We can exploit this characteristic by re-using the more generic features at the lower layers in a second round of training on a new subject or task. By freezing the lower layers, smaller datasets can be trained because it ruduces the number of network parameters that have to be optimized. This approach only has to learn features in the higher layers of the model, for previously unseen subjects or tasks. On the other hand, only when a large amount of data is available, a DL model can be successfully trained for specific users or tasks without transfering network parameters. Because it is very difficult to obtain sufficiently large amounts of data to train a model based on a specific user (i.e. in Standard learning or Distributed learning), the resulting models will perform worse than the models based on (\text{intra-experimental}) transfer learning (i.e. Split Learning or Frozen Learning).

With our work we show that a model is able to transfer some of its knowledge when performing a different classification task for the same paradigm (both motor imagery tasks). Extensions to the \textit{inter-experimental transferability} would be an investigation of the \textit{inter-paradigm transferability}. For example if a model that has been trained for a motor imagery classification task can be reused for a P300 detection task. An in-depth analysis of the transferability of knowledge between models trained on a different number of channels can also be an interesting extension. Such transferability could be done by freezing the middle layers of the model which would contain the core knowledge of a generalized model.

For the subject with the lowest performance all trained models often confuse the left and right hand with the movement of both feet. Perhaps EEG signals from this subject are not different enough to be distinguished due to the location of its dipoles or bridging between EEG electrodes placed over the primary motor cortex.

Finally, when exploring the transferability of models we see that the model is only able to outperform Standard Learning. This result is to be expected since using Distributed or Frozen Learning will take advantage of the large amount of data. It creates a model that is either tailored for the experiment or is tailored for the user in case of Frozen Learning.

The standard transfer technique is able to outperform Standard Learning, which means that it is possible to learn patterns by training a model on another experiment. This result is rather interesting when there is not a lot of data available, which makes it impossible to use Distributed or Frozen Learning.  


These results show that there should be an incentive for sharing models trained for EEG analysis. The models can be reused for different tasks by freezing specific layers. They can be shared to further train them on more data, thus improving the overall performance of the models and extending the number of possible applications in the field of BCI.




\section*{Bibliography}
\bibliographystyle{abbrv}
\bibliography{article}
\newpage

\end{document}